\documentclass[letterpaper, 10 pt, conference]{ieeeconf}  % Comment this line out if you need a4paper

\IEEEoverridecommandlockouts                              % This command is only needed if 
\usepackage{amsmath} % assumes amsmath package installed
\usepackage{amssymb}  % assumes amsmath package installed
\usepackage{graphicx}
\usepackage{subfigure}

\usepackage{xspace}
\makeatletter
\DeclareRobustCommand\onedot{\futurelet\@let@token\@onedot}
\def\@onedot{\ifx\@let@token.\else.\null\fi\xspace}
\def\eg{\emph{e.g}\onedot} 
\def\ie{\emph{i.e}\onedot}

\makeatother

\title{\LARGE \bf
Autonomous Stabilization of Retinal Videos for Streamlining Assessment of Spontaneous Venous Pulsations
}

\author{Hongwei Sheng$^{1}$$^{2}$, Xin Yu$^{2}$, Feiyu Wang$^{1}$, MD Wahiduzzaman Khan$^{1}$$^{2}$, Hexuan Weng$^{1}$,\\ Sahar Shariflou$^{1}$, S.Mojtaba Golzan$^{1}$% <-this % stops a space
\thanks{*This research is funded byARC-Discovery grant(DP220100800) and ARC-DECRA grant(DE230100477).}% <-this % stops a space
\thanks{$^{1}$Hongwei Sheng, MD Wahiduzzaman Khan, Hexuan Weng, Feiyu Wang, Sahar Shariflou and S.Mojtaba Golzan are with the University of Technology Sydney, 
        NSW 2007, Australia
        }
\thanks{$^{2}$Hongwei Sheng, MD Wahiduzzaman Khan, Xin Yu are with the University of Queensland,
        QLD 4067, Australia
        }%
}

\begin{document}

\maketitle
\thispagestyle{empty}
\pagestyle{empty}

%%%%%%%%%%%%%%%%%%%%%%%%%%%%%%%%%%%%%%%%%%%%%%%%%%%%%%%%%%%%%%%%%%%%
\begin{abstract} 
Spontaneous retinal Venous Pulsations (SVP) are rhythmic changes in the caliber of the central retinal vein and are observed in the optic disc region (ODR) of the retina. 
Its absence is a critical indicator of various ocular or neurological abnormalities. 
Recent advances in imaging technology have enabled the development of portable smartphone-based devices for observing the retina and assessment of SVPs. 
However, the quality of smartphone-based retinal videos is often poor due to noise and image jitting, which in return, can severely obstruct the observation of SVPs.
In this work, we developed a fully automated retinal video stabilization method that enables the examination of SVPs captured by various mobile devices.  
Specifically, we first propose an ODR Spatio-Temporal Localization (ODR-STL) module to localize visible ODR and remove noisy and jittering frames. 
Then, we introduce a Noise-Aware Template Matching (NATM) module to stabilize high-quality video segments at a fixed position in the field of view. 
After the processing, the SVPs can be easily observed in the stabilized videos, significantly facilitating user observations.
Furthermore, our method is cost-effective and has been tested in both subjective and objective evaluations. Both of the evaluations support its effectiveness in facilitating the observation of SVPs.
This can improve the timely diagnosis and treatment of associated diseases, making it a valuable tool for eye health professionals.
\end{abstract}

%%%%%%%%%%%%%%%%%%%%%%%%%%%%%%%%%%%%%%%%%%%%%%%%%%%%%%%%%%%%%%%%%%%%%%%%%%%%%%%%
\section{INTRODUCTION}

% Retinal vessel inspection is a widely used examination in clinical circumstance. By analyzing the shape and structure of the blood vessels in the retina via fundus photography, many physical characteristics associated with diseases can be clearly observed, thus enabling early identification of the condition.  However, manual methods for analyzing the retina vessels are time-consuming and overly rely on the expertise of medical professionals, especially on video format retina data. To fasten the large-scale analysis, many researchers are actively investigating the use of computer vision methods to automatically process retinal vessels. 

Spontaneous retinal venous pulsations (SVP) are rhythmic changes in the central retinal vein and its branches. SVPs are commonly present in the optic disc region (ODR) of the retina. Absent SVPs are clinically associated with progression in glaucoma and increased intracranial pressure \cite{understandingretinal,iih}. Accordingly, assessment of the retina in determining SVP presence is clinically paramount. 
%Performing regular checks of the status of SVP can greatly aid in the early detection of these diseases and avoid severe consequences. 

SVP evaluation is performed by inspecting the deformation of the retina vessels via fundus photography techniques \cite{atthepulsetime}. 
Conventional fundus photography data is usually captured using specialized and expensive benchtop equipment operated by trained professionals. However, with increasing emphasis on eye health, this method is not sufficient to meet the growing demand. 
% with the development of camera capabilities of smartphones, using smartphone-based  has become popular to 
% capturing fundus images with a smartphone \cite{} becomes popular as it is less expensive and more accessible. As a result

Recently, due to the low cost and easy accessibility of smartphones, researchers and clinicians \cite{wintergerst2020diabetic,iqbal2021smartphone} have commenced using smartphone-based fundus photography to assess retinal conditions, allowing frequent SVP observation.
However, retinal fundus images captured by hand-held devices are not robust to various real-world artifacts, such as noise and jittering. 
These artifacts would make medical analysis difficult and time-consuming \cite{MONJUR2021100177,affect,xin}, thus presenting challenges in clinical diagnosis and exacerbating the likelihood of erroneous evaluation.
% greatly by real-world issues, such as inexperienced operators/photographers, unpredictable eyeball movements and inevitable blinks.
% unpredictable environments, and uncooperative patients.% inadequate equipment,
% More importantly, it also highly relies on the skill of the medical professional, thus restricting its usage in practice.

% Seeking the field of video stabilization, many good
Most existing video stabilization works are proposed to stabilize natural scenes \cite{GUILLUY2021116015} without taking specific image domain knowledge into account. 
% are based on videos of human motion or other natural scenes . 
Only a few works \cite{9175461,mueller2020automated} have been proposed to stabilize fundus retina videos. 
However, those methods often require high-quality videos. For example, the videos cannot contain eye blinks and drastic illumination changes. Thus, those methods are not suitable to process mobile-captured real-world retina videos.
Efficiently monitoring SVPs with computer-assisted techniques \cite{mchugh2020spontaneous,insufficient} still remains challenging. It is necessary to design an effective automatic mobile-based fundus video stabilization method to enable easy observation and diagnosis.
% complete autonomy from physician involvement
% Thus, 
% Even with the use of  
% To enhance the current situation and reduce the workload of the healthcare system, 
\begin{figure}[t]
    \centering
    \includegraphics[width=0.95\linewidth]{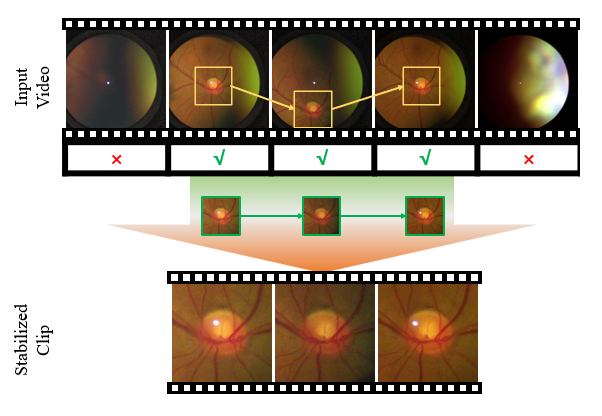}
        \vspace{-1em}
    \caption{Illustration of our proposed automatic fundus retina video stabilization method. Given a fundus retina video, our method automatically detects the ODR both spatially and temporally. 
    Afterward, our approach aligns the visible ODR clips to a fixed position in the video footage. The stabilized video facilitates better SVP inspection.}
    \label{fig:brief}
    \vspace{-1em}
\end{figure}

% \begin{figure}[t]
% \begin{center}
% \includegraphics[width=0.95\linewidth]{}
%    \caption{}
% \label{fig-1}
% \end{center}
% \vspace{-2em}
% \end{figure}

\begin{figure*}[ht]
    \centering
    \includegraphics[width=0.9\linewidth]{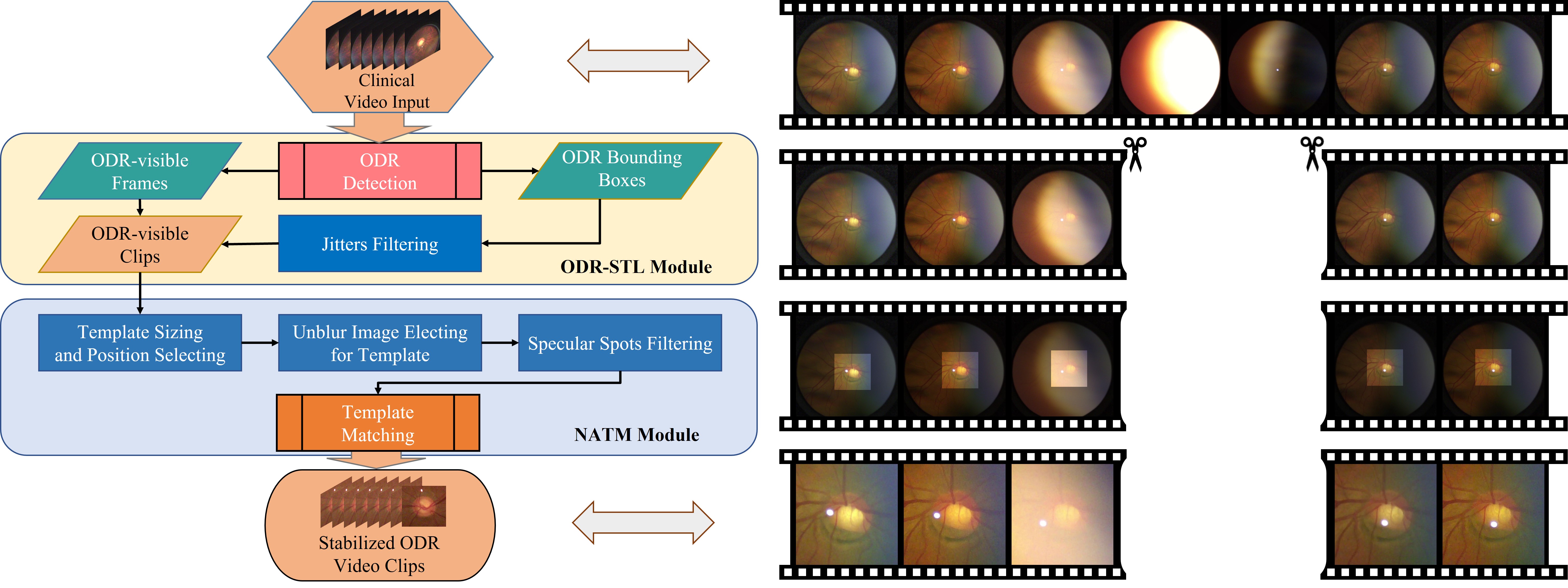}
        \vspace{-1em}
    \caption{Overview of the pipeline of our method.
    Left: The video will be first processed by an ODR Spatio-Temporal Localization Module to remove ODR-invisible frames. Then, a Noise Aware Template Matching is applied to align high-quality ODR frames to a fixed position. The video is cropped to focus on ODR regions, thus aiding SVP inspection. 
    Right: The intermediate results of our processing pipeline.}
    \label{fig:pipeline}
    \vspace{-1em}
\end{figure*}

To better support clinicians or even non-experts in observing and evaluating SVPs, we propose a fully automatic fundus retina video stabilization method as shown in Fig. \ref{fig:brief}.
% Specifically, our method first detects the ODR from a fundus video and then localizes high-quality video frames in which ODR can be clearly observed in consecutive frames (more than 1 second). 

Specifically, we design an ODR Spatio-Temporal Localization (ODR-STL) module to first detect the spatial locations of ODR from a fundus video and then temporally localize the ODR-visible clips from the video.

Next, we introduce a Noise-Aware Template Matching (NATM) module to stabilize the ODR-visible video clips to a fixed position of the footage (\ie, the field of view). In the meanwhile, NATM is robust to specular near ODR regions since specular might severely affect conventional template matching performance.
% are then stabilized by aligning the ODR at a fixed position in the footage. 

After the stabilization, the observation of ODR videos will not be subject to the artifacts caused by eyeball movements, blinks, or abrupt brightness fluctuation, thus significantly facilitating clinical assessment and downstream machine perception.
To further illustrate the effectiveness of our proposed method, we apply it to real fundus videos captured with various types of smartphone-based fundus cameras. Moreover, we quantitatively evaluate the quality of the stabilized video clips as well as conduct a user study in which clinicians are invited to choose which videos are more favored in observing and evaluating SVPs.

\section{METHODOLOGY}

% \subsection{Task Definition}
Our fully automated stabilization method is designed to facilitate the detection of SVPs from clinical fundus videos. 
As shown in Fig. \ref{fig:pipeline}, we feed an input fundus retina video captured by the smartphone-based retina imaging platform in clinics to our system and then obtain a stabilized video of optical disc regions, where SVPs usually emerge.
% with a resolution of 1800$\times$1800.
In our method, the first step is a spatio-temporal localization of visible ODR (ODR-STL) and then we select continuous ODR-visible frames. 
The second step is to apply noise-aware template matching (NATM) to stabilize the positions of visible ODR in video clips. 
\subsection{Spatio-temporal Localization of Visible ODR}
 
% This module is the first step in pre-processing. 
% The first step in our method is to localize ODR-visible frames in a video on both spatial dimension and temporal dimension.
% Moreover, jitters in the footage can lead to a blurry image while continuous jitters may obstruct the assessment of SVPs.  
%Too high in frequency or too violent in amplitude
% To address these problems, we design
% We then design a two-step approach to achieve 
To accurately determine the presence of SVP, clinicians need to observe at least one full clear cycle of the pulsation in a fundus video.
Thus, our ODR-STL method first detects frames that contain ODR in the footage.
% This provides ODR positions in frames spatially and ODR visibility through the frames temporally.
% % The spatial localization of faster R-CNN provides positions of ODR in frames and the temporal 
% By calculating the ODR positions over frames, our algorithm then identifies huge jitters and eliminates them.
Then our algorithm identifies and eliminates huge jitterings in the footage.
%  % re-spliced 
As a result, video clips that contain continuous ODR with sufficient temporal duration (\ie, longer than a full cycle of SVP) are kept for further stabilization.

\subsubsection{ODR Detection} 
Our method first employs a faster R-CNN neural network \cite{fasterrcnn} to detect ODR in video frames.
% A faster R-CNN neural network \cite{fasterrcnn} was trained to detect ODR in video frames. 
Utilizing deep networks helps to improve the accuracy of the detecting results and circumvents some limitations of traditional methods (\eg, sensitive to noise and illumination changes).
% The trained Faster R-CNN model gives out the timestamps (\emph{i.e}, the frame No) of when the ODR appears in the temporal sequence as well as the bounding box of ODR in each frame. 
With the faster R-CNN model, our method extracts the bounding box of ODR in each frame as well as obtains the existence of ODR along the temporal dimension.
% timestamps (\emph{i.e}, the frame No) of when the ODR appears in the temporal sequence.
% This provides ODR positions in frames spatially and ODR visibility through the frames temporally.
As a result, we obtain the spatial and temporal information of visual ODR in a video.
% With the temporal timestamp information, the videos can be split into one or more clips by removing the frames with no visible ODR. 
% We also manually labeled the timestamps and compared them with the automatically generated timestamps. 
% The mean Intersection over Union (IOU) was 96.98.

\subsubsection{Filtering Jitters} 
\begin{figure*}[t]
  \begin{minipage}[b]{0.35\textwidth}
    \centering
    \scalebox{1.}[1]{\includegraphics[width=\textwidth]{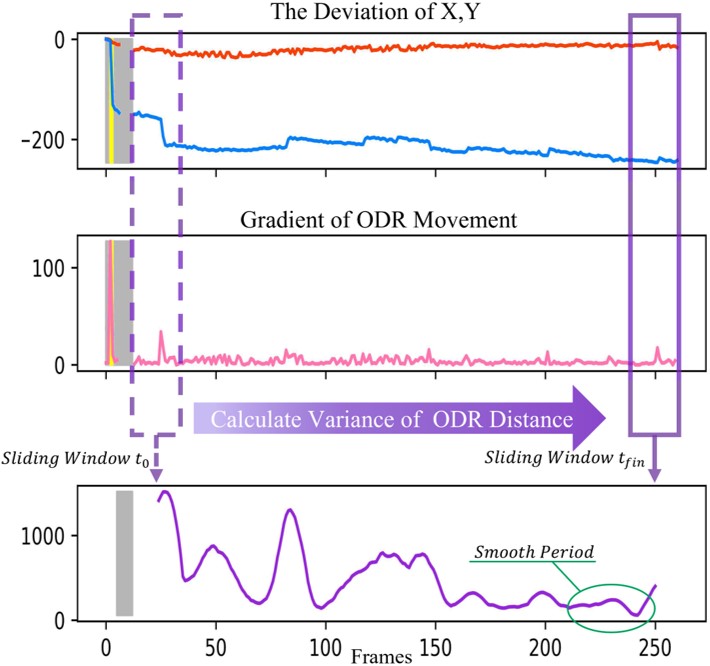}}
        \vspace{-1.5em}
    \caption{ODR Trajectories. The red (X-axis) and blue (Y-axis) lines indicate detected ODR deviated from the position in the first frame. 
    % The detected ODR in the first frame is used as baseline and the deviations in the following frames are shown. 
    % The red one indicates the deviations along the x-axis and the blue one represents the deviations along the y-axis. Bottom: The green line reflects the gradient of detected ODR trajectories. 
    The gray regions indicate the removed frames after localization. The pink line indicates the gradient of ODR trajectories. The purple line indicates the variance of ODR distance over frames.}
    % Both start from the position of the first detected ODR.
    \label{local}
        % \vspace{-1em}
  \end{minipage}
 \hfill
  \begin{minipage}[b]{0.6\textwidth}
    \centering
    % \scalebox{0.92}[0.87]{\includegraphics[width=\textwidth]{template/exps/demo.png}}
    \scalebox{1}[1]{\includegraphics[width=\textwidth]{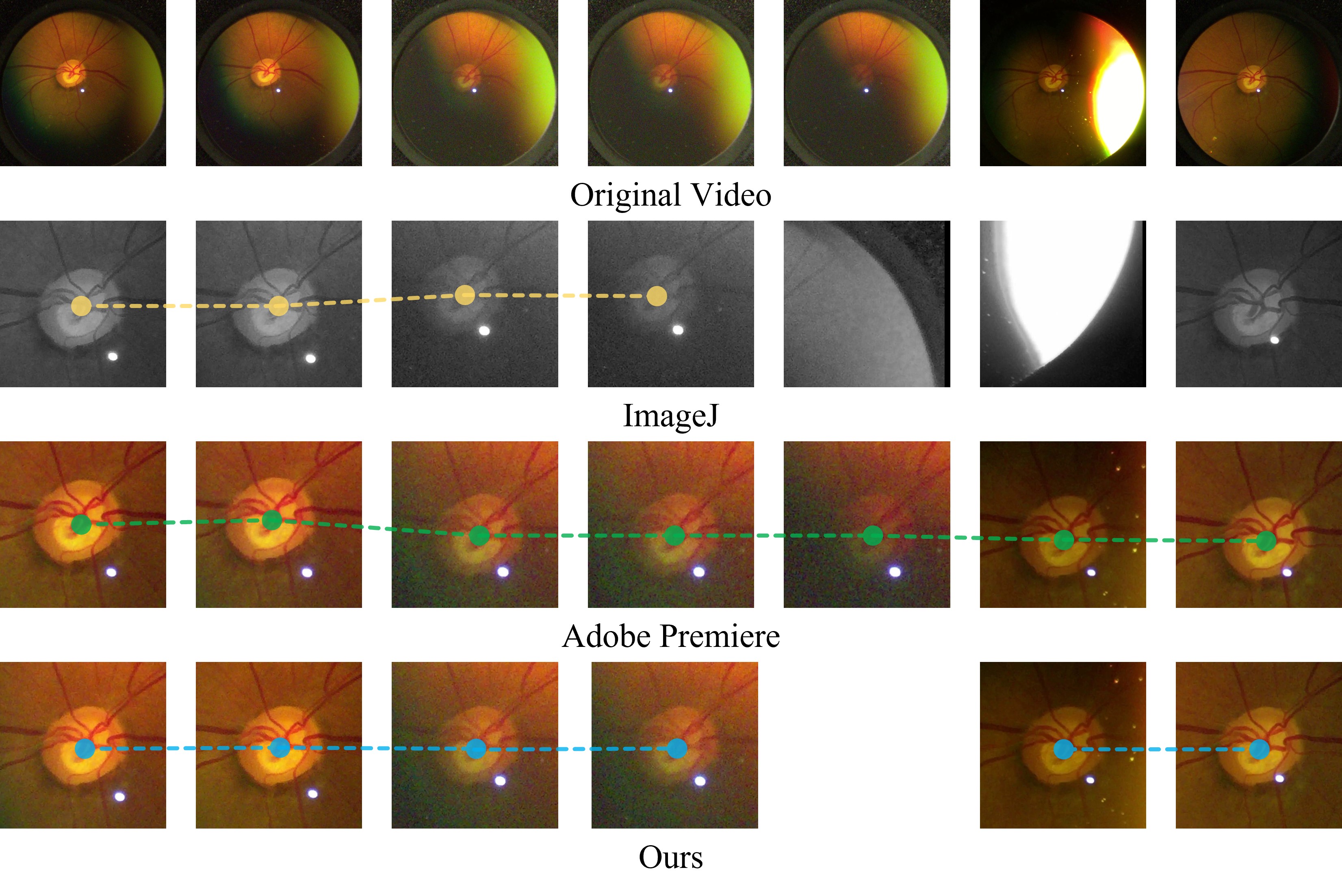}}
        \vspace{-1.8em}
    \caption{Comparison with other video stabilization methods. Note that our method is able to remove low-quality images. In this example, ODR does not have enough illumination and thus we remove this frame and obtain two video clips.}
    \label{fig:compare}
        \vspace{-5.2em}
  \end{minipage}
  \vspace{-1.1em}
  \end{figure*}

% Our method then further proceeds with spatial localization by utilizing bounding boxes from faster R-CNN results.
% The detection results of faster R-CNN also contain . 
% The bounding boxes  
Our method then utilizes bounding boxes to eliminate the jitters in the footage.
Due to the inherent noise presented in the video data and the size variance of detected boxes, the detected bounding boxes present frequent fluctuations. 
% Though the jittering makes it impractical to use these bounding boxes directly for clinical purposes, 
However, since the fluctuation around the ODR is small compared to the sizes of bounding boxes, our method still obtains the approximate location of the ODR in each frame, enabling further screening extreme jitters.
As shown in Fig. \ref{local}, the polylines reflect the movements of ODR in the footage throughout the video. 
The broken parts of the polyline indicate that the ODR is not detected in these frames.
By eliminating the huge jitter marked as yellow vertical, our method further preserves a continuous clip with ODR consistently showing in footage.

\subsection{Noise-Aware Template Matching}

% This module is the second stage of SVP stabilization. 
% Its nature is to use traditional computer vision algorithms such as Template Matching \cite{} to register the position of ODR in every frame. 
% To achieve this, video clips showing unobscured ODR is required (i.e., the aforementioned stage). 
Our NATM method is then applied to register the positions of ODR in each clip.
Due to the presence of random noise, it is unsuitable to apply the plain template matching algorithm to the data.
Therefore, three additional strategies have been introduced to mitigate the impact of noise when applying template matching. 
% Tradition computer vision algorithm Template matching is an ideal choice.
% After temporal localization from the first part,
%Optic flow variant -> Optical flow variations or magnitudue of optical flow between consecutive frames
 
%Filtering high RGB -> Image specular spots removal. The specular spots are caused by eye balls’ specular reflection.

\begin{figure}
    \centering
    \includegraphics[width=0.8\linewidth]{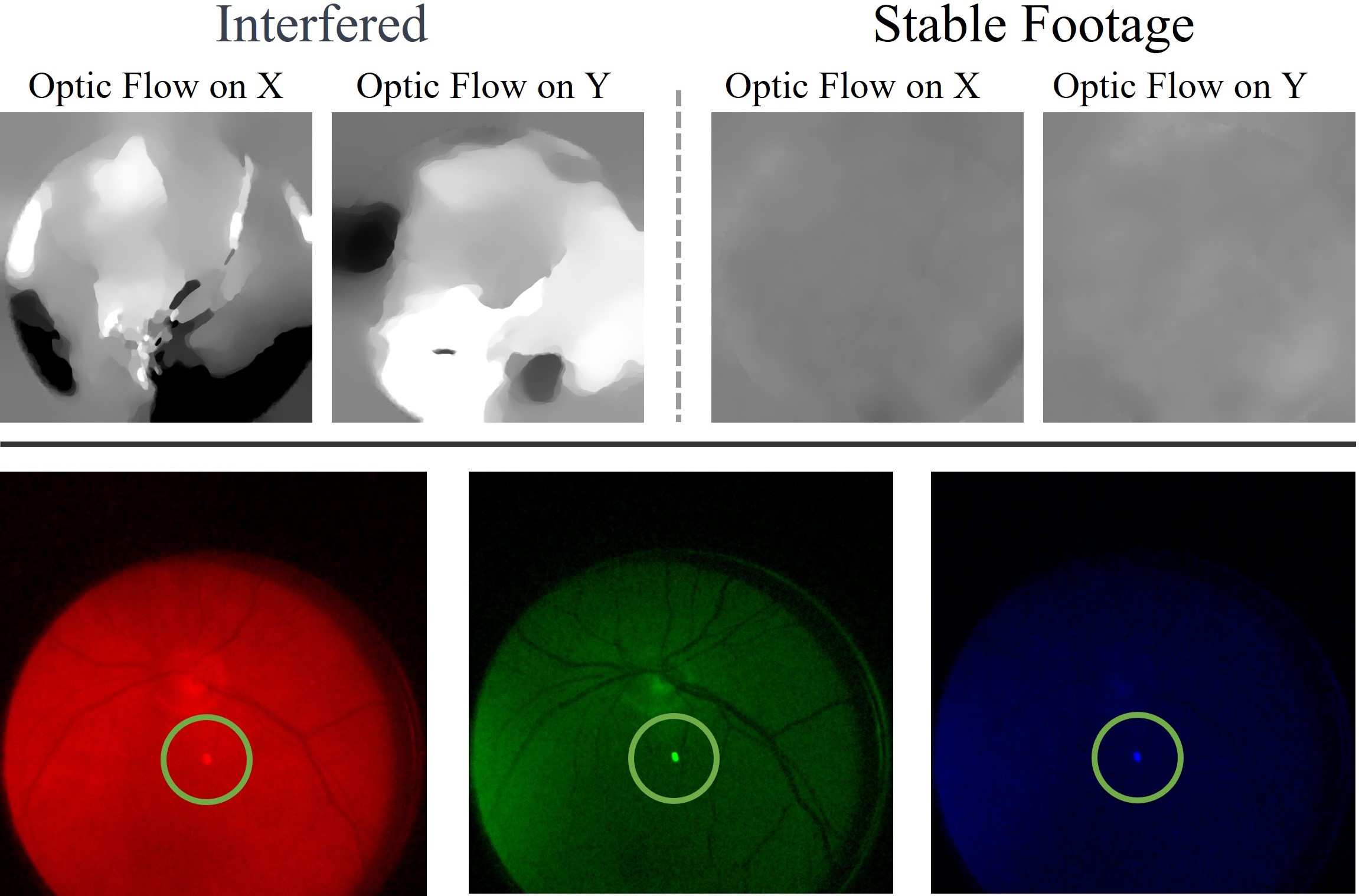}
    % \vspace{-1em}
    \caption{Top: Optical flow before and after stabilization. Top left: the optical flow of an original video along X and Y dimensions, respectively. Top right: the optical flow of our stabilized video. The smoother flow indicates that the video has fewer jittering artifacts. Bottom: Impacts of specular on different channels in template matching. We can observe that the blue and green channels are very sensitive to specular. Therefore, in order to achieve a noise-aware template matching, our method reduces the impacts of blue and green channels on template matching.}
    \vspace{-1.5em}
  %   % Top: The polylines indicate the number of pixels along their values in the corresponding channel. The color of the polyline refers to the color of the channel. Bottom: The frame with this distribution. This frame contains specular spots and other illumination obstructions.
  %   % Optical flow before and after stabilization. Top: the optical flow of an original video along X and Y dimensions, respectively. Bottom: the optical flow of our stabilized video. The smoother flow indicates that the video has fewer jittering artifacts.}
    \label{histo}
\end{figure}

\subsubsection{Template Size Selection} 
Template-based matching requires a template as a reference. 
The algorithm identifies the region of interest by evaluating the pixel-wise difference in RGB values between a template and a potential matching target. 
Thus, it is important to choose a template from a region that is significantly different from the other parts of the image.
To achieve the best matching results on clinical noisy data, the size and content of the template are very crucial.
Since ODR is the brightest area and is unique in retina images, the best choice for our task is to employ a rectangle template centered on the ODR, with the side length overlapping the ODR diameter. 
The ODR diameter is estimated from the sizes of bounding boxes detected by the faster R-CNN.
% Since the faster R-CNN is trained to fit the tightest results, it will return the same results for the same inputs.
% Using the results from the bounding boxes detected by the faster R-CNN in the first module, a large number of measurements for the ODR size can be obtained. 
% By setting a threshold for these sizes, the algorithm can automatically select the appropriate side length for the template and use it for matching.

\subsubsection{Screening Blur Template}
After confirming the size and position of the template, our method then determines which frame should be adopted as the template.
In order to accurately find the ODR, our method selects the template from a video frame without blur. 
Considering the irregular noise in a video, it is difficult to calculate sharpness via conventional gradient-based algorithms such as Laplacian sharpness measurement. 
% Meanwhile, we notice blur frames usually occur when jitters are present. 
Meanwhile, we notice that blur frames usually enlarge the jitters of bounding boxes.
Therefore, in each video clip, we use the trajectory of bounding box centers within a sliding window to select the most smooth period of the video, as shown in Fig. \ref{local}. 
Then, we use the variance of optical flow \cite{optic} to determine the quality of each frame in terms of sharpness as shown in Fig. \ref{histo}. 
The frame with the lowest variance of the optic flow will be considered as the sharpest image in the most smooth period of the clip.
% the optical flow variants of frame-wise in the window and then returns a frame with the lowest variant of itself. We consider this frame is of the sharpest image in the most steady period.
% Template-based matching requires a reference template. To enhance the accuracy in finding the ODR, it is essential to select a template with no blur. Due to the presence of irregular noise, it is challenging to achieve sharpness using conventional gradient-based algorithms such as Laplacian sharpness calculation. Additionally, jittering caused by either the smartphone or the ODR can result in blurred frames with a high probability. To address this, we use optical flow data to determine the clarity of each frame. A sliding window is employed to calculate the optical flow variations of frames within the window and the frame with the lowest variation is selected as the sharpest image in the most stable period.
%  based on optical flow \cite{}
%a high-resolution image with high sharpness should be our ideal choice as a template
\subsubsection{Specular Spots Removing} 
In practice, some hand-held fundus video-capturing devices utilize external light sources to better observe fundus retina. 
This results in specular spots on the eyeballs due to the reflection. 
As the light sources normally emit white light, the reflected specular spots present high values in RGB channels as shown in Fig. \ref{histo}. 
% Since the human eyeballs are translucent and should be orange-yellow in imaging, it is abnormal for the pixels to contain higher values in green and blue channels than in red channels. 
We can observe that specular spots protrude especially in the blue (B) and green (G) channels.
Thus, we selected an appropriate global threshold on the B and G channels. 
% Thus, we counted the RGB value distribution of each frame of a video and selected an appropriate global threshold on the B and G channels. 
Then, our algorithm employs mean filtering to minimize the interference of specular light spots during template matching.

With these three strategies, template matching obtains a series of ODR coordinates precisely from each noisy video clip.
This allows us to easily align these coordinates and fix the positions of ODR in the footage. 
Our method further crops the ODR out with a modifiable size to emphasize SVP.
As a result, we obtain a set of ODR-stabilized video clips from our auto-processing pipeline.

% This allows the algorithm to easily crop and highlight the ODR in the video, as per our requirements. 
% In our study, we cropped 640$\times$640 clipped videos from the original 1800$\times$1800 full-length videos. 
% This not only reduces the video's size but also eliminates any unnecessary information, thereby emphasizing the ODR and the area where SVPs are commonly detectable. 
% This allows clinicians to observe and assess SVPs.
% %Calculating Optic flow variant to select the most steady frames as template bases in clips; Filtering high RGB to avoid the affection of white light speckles; Select the most complex part of disc region by narrowing down the kernel size of template matching to increase the gap of potential matching templates/windows and dodge large light spots.

% The template is used to format your paper and style the text. All margins, column widths, line spaces, and text fonts are prescribed; please do not alter them. You may note peculiarities. For example, the head margin in this template measures proportionately more than is customary. This measurement and others are deliberate, using specifications that anticipate your paper as one part of the entire proceedings, and not as an independent document. Please do not revise any of the current designations
\begin{figure*}
    \centering
    % \scalebox{1}[0.8]{\includegraphics[width=0.8\linewidth]{template/exps/keyi.png}}
    \scalebox{1}[1]{\includegraphics[width=\linewidth]{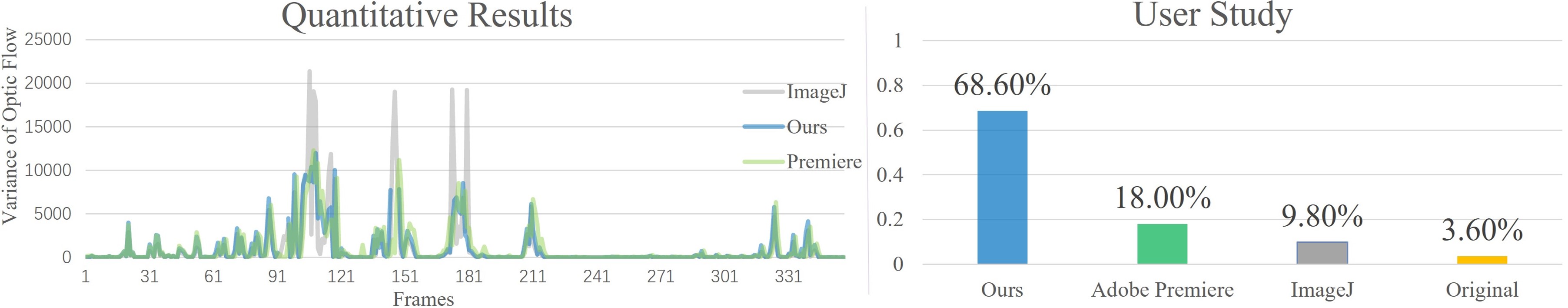}}
    \vspace{-1.67em}
    % \caption{User Study. There are four different categories standing for different preferences of 25 subjects for videos from four sources.}
    \caption{Left: The variance of optic flow over frames. The variance of our method is the lowest. The variance of Adobe Premiere is close to ours, but its effort on reducing optic flow variance brings distortion to the stabilized video. Right: User Study. We invite 25 subjects including clinical professionals and non-experts to evaluate the quality of 20 stabilized videos by different methods. The original videos are also provided to the users for evaluation. Most users favor stabilized videos by our method.}
    \label{fig:case}
    \vspace{-0.8em}
\end{figure*}

\section{EXPERIMENTS}

In our study, our pipeline will automatically crop video clips of 640$\times$640 pixels from the original videos of size 1800$\times$1800 pixels. 
This not only saves the storage by reducing redundant information, but also emphasizes the area where SVPs are commonly detectable. 
We then produce quantitative and qualitative experiments including a user study to evaluate the performance of our fundus retina video stabilization method against some popularly used approaches.
\subsection{Quantitative Results}
% 1. method -->
% 2. quantitative results. 

To further illustrate the effectiveness of our method in inspecting SVPs, we conduct a series of comparative experiments as objective evaluations.
We also compare with some existing video stabilization methods.
ImageJ (\ie, FIJI) 
% \cite{schindelin2012fiji} 
% \cite{schneider2012nihimj}
is a widely used software in medical image processing. 
Its plugins provide capacity for many tasks including video stabilization.
Adobe Premiere Pro is a popular video processing and editing commercial software. 
It has a built-in function to stabilize videos as well.
% These methods are applied to stabilize the original fundus videos and their stabilized results are used to be compared with our method.
% For equality, we crop the videos stabilized by other methods to the same size as our method.
We utilize optic flow to measure the stability.
As shown in Fig. \ref{fig:case}, our method achieves the least variance of optical flow.

\subsection{Qualitative Results}

Fig. \ref{fig:compare} presents examples from our method and other methods.
The original video of this example contains blur, specular spots, low illumination, and other real-world noise. 
We can see that ImageJ fails to address the blur and the stabilized videos suffer jittering when the specular moves. 
In the video clips with less noise, the performance of ImageJ can be on par with our method.
Adobe Premiere Pro performs more steadily, but it introduces distortion in the stabilized video. This distortion will harm the SVP observation. 

% \begin{figure}
%     \centering
%     \includegraphics[width=\linewidth]{template/pics/compare.jpg}
%     \caption{(draft)The Comparison between Our Method and Others }
%     \label{fig:compare}
% \end{figure}

\subsection{User Study}
% To access the 

To determine the effectiveness of our method for manual diagnosis, we recruited 25 individual subjects with varying levels of expertise from four different clinics. 
We prepare some groups of test videos in advance. 
Each group contains four videos processed by four different methods from the same original video.
We then ask the subjects to watch five groups of test videos and then report which video in each group they would like to use to best observe SVP.
We graph their feedback as in Fig. \ref{fig:case}.
In this user study, videos from our method are more favored than those from the others.
% Each subject w It supports video stabilization with template-matching plugins and has similar functionality as oursas provided with a pair of visible SVP videos and a pair of invisible SVP videos in each experiment.
% Considering variable control, we conducted three experiments with different video sizes.

% In the first experiment, both videos of each pair were kept unchanged (with sizes of 1800$\times$1800 and 640$\times$640 respectively) and subjects were asked to view the videos at the same observation distance. 
% 90\% of the subjects felt that the processed video was clearer. 
% In the second experiment, the stabilized video was scaled up to the same size as the original video (1800$\times$1800) and the viewing distance was kept constant. Only 57.5\% of the subjects felt that our processing was more effective while 20\% felt there was no significant difference. 
% % This is partly due to the limited receptive field of the human eye at a fixed distance, and also because the scaling-up algorithm in computer vision amplifies the noise in the video, making it more difficult for subjects to observe. 
% In the third experiment, the original video was scaled down to the same size as our processed video. 
% The subjects unanimously reported that the original video is not capable to compete with the processed video at this time.

% Among these three experiments, two of them show that our processed videos have greatly improved the visibility of SVP compared to the original videos.
% Our stabilized videos show bad capacity in the second 

\section{CONCLUSIONS}

% In this paper, we propose an automatic preprocessing method that streamlines SVPs assessment in a clinical setting by stabilizing ODR in a fixed position throughout the video. 
% Our method also externalizes the robust usability of its stabilization. 
% Our user study and experiments ultimately demonstrate the effectiveness of our method and illustrate the potential benefits our method can bring to further analysis. 
% Besides, our method can be easily adapted to different tasks with other points of interest by calculating the relative positions of the ODR to other regions of interest.

In this paper, we propose an automatic fundus retina video stabilization method that would profoundly promote the wide application of smartphone-based SVP inspection and assessment. From the stabilized videos, clinical professionals or even non-professionals can easily observe SVPs. We also believe the stabilized video can significantly ease the downstream clinical diagnosis and analysis. Noticeably, our method has been tested on various data collected by different mobile devices in different clinics. The experiments demonstrate that our retina video stabilization method is very effective in different clinical environments.
%  to conduct user studies and experiments.
\addtolength{\textheight}{-12cm}   % This command serves to balance the column lengths
                                  % on the last page of the document manually. It shortens
                                  % the textheight of the last page by a suitable amount.
                                  % This command does not take effect until the next page
                                  % so it should come on the page before the last. Make
                                  % sure that you do not shorten the textheight too much.

%%%%%%%%%%%%%%%%%%%%%%%%%%%%%%%%%%%%%%%%%%%%%%%%%%%%%%%%%%%%%%%%%%%%%%%%%%%%%%%%

%%%%%%%%%%%%%%%%%%%%%%%%%%%%%%%%%%%%%%%%%%%%%%%%%%%%%%%%%%%%%%%%%%%%%%%%%%%%%%%%

%%%%%%%%%%%%%%%%%%%%%%%%%%%%%%%%%%%%%%%%%%%%%%%%%%%%%%%%%%%%%%%%%%%%%%%%%%%%%%%%
% \section*{APPENDIX}

% Appendixes should appear before the acknowledgment.

% \section*{ACKNOWLEDGMENT}

% The preferred spelling of the word ``acknowledgment'' in America is without an ``e'' after the ``g''. Avoid the stilted expression, ``One of us (R. B. G.) thanks . . .''  Instead, try ``R. B. G. thanks''. Put sponsor acknowledgments in the unnumbered footnote on the first page.
\vspace{-0.4em}
\bibliographystyle{ieeetr}
\bibliography{main.bib}
% \vspace{0.7em}
% \clubpenalty 10000

\end{document}